# Isotropic reconstruction of 3D fluorescence microscopy images using convolutional neural networks

Martin Weigert[1,2], Loic Royer[1,2], Florian Jug[1,2], and Gene Myers[1,2,3]

[1] Max Planck Institute of Molecular Cell Biology and Genetics, Dresden, Germany
[2] Center for Systems Biology Dresden, Germany
[3] Faculty of Computer Science, Technical University Dresden, Germany

**Abstract.** Fluorescence microscopy images usually show severe anisotropy in axial versus lateral resolution. This hampers downstream processing, i.e. the automatic extraction of quantitative biological data. While deconvolution methods and other techniques to address this problem exist, they are either time consuming to apply or limited in their ability to remove anisotropy. We propose a method to recover isotropic resolution from readily acquired anisotropic data. We achieve this using a convolutional neural network that is trained end-to-end from the same anisotropic body of data we later apply the network to. The network effectively learns to restore the full isotropic resolution by restoring the image under a trained, sample specific image prior. We apply our method to 3 synthetic and 3 real datasets and show that our results improve on results from deconvolution and state-of-the-art superresolution techniques. Finally, we demonstrate that a standard 3D segmentation pipeline performs on the output of our network with comparable accuracy as on the full isotropic data.

## 1 Introduction

Fluorescence microscopy is a standard tool for imaging biological samples [16]. Acquired images of confocal microscopes [4] as well as light-sheet microscopes [5], however, are inherently anisotropic, owing to a 3D optical point-spread function (PSF) that is elongated along the axial (z) direction which typical leads to a 2 to 4-fold lower resolution along this axis. Furthermore, due to the mechanical plane-by-plane acquisition modality of most microscopes, the axial sampling is reduced as well, further reducing the overall resolution by a factor of 4 to 8. These effects later render downstream data analysis, e.g. cell segmentation, difficult.

To circumvent this problem, multiple techniques are known and used: Classical deconvolution methods [13, 10] are arguably the most common of these. They can be applied on already acquired data, however, their performance is typically inferior to other more complex techniques. Some confocal systems, e.g. when using two-photon excitation with high numerical aperture objectives and an isotropic axial sampling, can acquire almost isotropic volumes [4, 11] (cf. Figure 3). Downsides are low acquisition speed, high photo toxicity/bleaching, and large file sizes. Light-sheet microscopes, instead, can improve axial resolution by imaging the sample from multiple sides (views). These views can then be registered and jointly deconvolved [12]. The disadvantage is the reduced effective acquisition speed and the need for a complex optical setup. A method that would allow to recover isotropic resolution from a single, anisotropic acquired microscopic 3D volume is therefore highly desirable and would likely impact the life-sciences in fundamental ways.

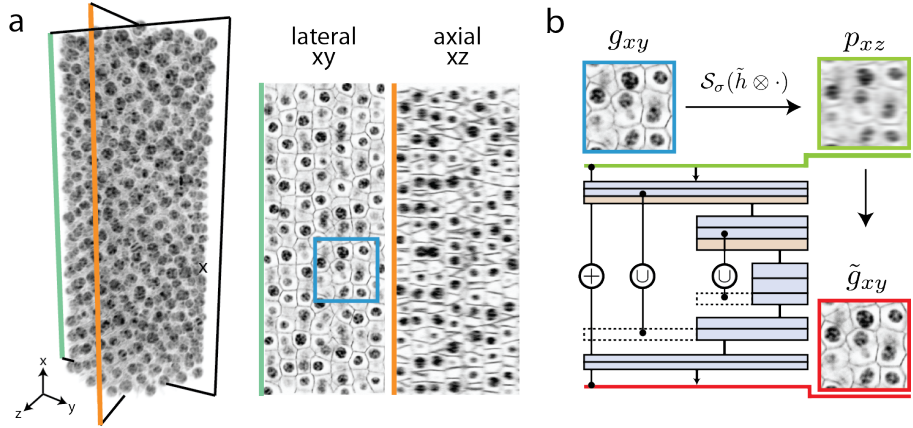

Fig. 1: a) 3D images acquired on light microscopes are notoriously anisotropic due to axial undersampling and optical point spread function (PSF) anisotropy. b) The IsoNet-2 architecture has a U-net [14] like topology and is trained to restore anisotropically blurred/downsampled lateral patches. After training it is applied to the axial views.

Here we propose a method to restore isotropic image volumes from anisotropic light-optical acquisitions with the help of convolutional networks without the need for additional ground truth training data. This can be understood as a combination of a super-resolution problem on subsampled data, and a deconvolution problem to counteract the microscope induced optical PSF. Our method takes two things into account: ($i$) the 3D image formation process in fluorescence microscopes, and ($ii$) the 3D structure of the optical PSF. We use and compare two convolutional network architectures that are trained end-to-end from the same anisotropic body of data we later apply the network to. During training, the network effectively learns a sample specific image prior it uses to deconvolve the images and restore full isotropic resolution.

Recently, neural networks have been shown to achieve remarkable results for super-resolution and image restoration on 2D natural images where sufficient ground truth data is available [3, 2, 7]. For fluorescence microscopy data there is, unfortunately, no ground truth (GT) data available because it would essentially require to build an ideal and physically impossible microscope. Currently there is no network approach for recovering isotropic resolution from fluorescence microscopy images. Our work uses familiar network architectures [14, 1], and then applies the concept of *self super-resolution* [6] by learning from the very same dataset for which we restore isotropic resolution.

## 2 Methods

Given a true fluorophore distribution $f(x, y, z)$ the acquired volumetric image $g$ of a microscope can be approximated by the following process

$$g = \mathcal{P}\big[\mathcal{S}_\sigma(h \otimes f)\big] + \eta \qquad (1)$$

where $h = h(x, y, z)$ is the point spread function (PSF) of the microscope, $\otimes$ is the 3D convolution operation, $\mathcal{S}_\sigma$ is the axial downsampling/slicing operator by a factor $\sigma$, $\mathcal{P}$ is the signal dependent noise operator (e.g. poisson noise) and $\eta$ is the detector noise. As the PSF is typically elongated along $z$ and $\sigma > 1$, the lateral slices $g_{xy}$ of the resulting volumetric images show a significant higher resolution and structural contrast compared to the axial slices $g_{xz}$ and $g_{yz}$ (cf. Figure 1a).

### 2.1 Restoration via convolutional neural networks

The predominant approach to invert the image formation process (1) is, in cases where it is possible, to acquire multiple viewing angles of the sample, and register and deconvolve these images by iterative methods without any sample specific image priors [10, 13, 12]. In contrast to these classical methods for image restoration, we here try to directly learn the mapping between blurred and downsampled images and its true underlying signal. As no ground truth for the true signal is available, we make use of the resolution anisotropy between lateral and axial slices and aim to restore lateral resolution along the axial direction. To this end, we apply an adapted version of the image formation model (1) to the lateral slices $g_{xy}$ of a given volumetric image

$$p_{xy} = \mathcal{S}_\sigma(\tilde{h} \otimes g_{xy}) \qquad (2)$$

with a suitable chosen *rotated PSF* $\tilde{h}$. To learn the inverse mapping $p_{xy} \to g_{xy}$ we assemble lateral patches $(g_{xy,n}, p_{xy,n})_n$ and train a fully connected convolutional neural network [9] to minimize the pixel wise PSNR loss

$$\mathcal{L} = \sum_n -[20\log_{10} \max g_{xy,n} - 10\log_{10} |g_{xy,n} - \tilde{g}_{xy,n}|^2] \qquad (3)$$

where $\tilde{g}_{xy,n}$ is the output of the network when applied to $p_{xy,n}$. For choosing the best $\tilde{h}$ we consider the two choices (i) *full*: $\tilde{h} = h_{rot}$ where $h_{rot}$ is a rotated version of the original PSF that is aligned with the lateral planes, and (ii) *split*: $\tilde{h} = h_{split}$ which is the solution to the deconvolution problem $h_{rot} = h_{iso} \otimes h_{split}$ and $h_{iso}$ is the isotropic average of $h$. The later choice is motivated by the observation that convolving lateral slices with $h_{split}$ leads to images with a resolution comparable to the axially ones. After training we apply the network on the unseen, anisotropically blurred, bicubic upsampled axial slices $g_{xz}$ of the whole volume to get the final estimation output.

### 2.2 Network architecture and training

We propose and compare two learning strategies, *IsoNet-1* and *IsoNet-2* , which are implementing two different established network topologies. The notation for

the specific layers is as follows:

$C_{n,w,h}$ Convolutional layer with $n$ filters of size $(w,h)$
$M_{p,q}$ Max pooling layer with a subsample factor of $(p,q)$
$U_{p,q}$ Upsampling layer with a subsample factor of $(p,q)$

In conjunction with the two different methods of training data generation (*full*, *split*), the specific topologies are:

*Isonet-1* Which is the proposed network architecture of [1] used for super resolution:
$$C_{64,9,9} - C_{32,5,5} - C_{1,5,5} - C_{1,1,1}$$
Here the first layer acts as a feature extractor whose output is mapped non-linearly to the resulting image estimate by the subsequent layers. After each convolutional layer a rectifying activation function (ReLu) is applied.

*Isonet-2* Which is similar to the proposed network architecture of [14] for segmentation which consists of a contractive part
$$C_{16,7,7} - M_{2,2} - C_{32,7,7} - M_{2,2} - C_{64,7,7} - U_{2,2} - C_{32,7,7} - U_{2,2} - C_{16,7,7} - C_{1,1,1}$$
that learns sparse representations of the input and skip connections that are sensitive to image details (cf. Figure 1b). In contrast to [14], however, the network learns the residual to the input. ReLu activation is used throughout.

For all datasets, both architectures were trained for 100 epochs with the Adam optimizer [8] and a learning rate $5 \cdot 10^{-3}$. We furthermore use a dropout of 20% throughout and apply data augmentation (flipped and rotated images) where it is compatible with the symmetries of the PSF.

## 3 Results

### 3.1 Synthetic Data

We use 3 synthetic datasets, as shown in Figure 2. The uppermost row shows small axial crops from a volume containing about 1500 simulated nuclei. The middle row shows crops of membrane structures as they are frequently seen in tightly packed cell epithelia. The third and last row shows both, simulated cell nuclei and surrounding labeled membranes. Note that the first column shows the ground truth images that were used to generate the isotropic ground truth, by convolving with the isotropic PSF, and the blurred images that were subsampled and convolved with realistic PSFs in order to resemble microscopic data. This third column (blurred) is then used as the input to all our and other tested methods. The subsequent 6 columns show the results of ($i$) Richardson-Lucy deconvolution [10], ($ii$) pure SRCNN [1], i.e. disregarding the PSF, ($iii$) the IsoNet-1 using the full PSF, ($iv$) the IsoNet-1 using the anisotropic component of the PSF $h_{split}$, ($v$) the IsoNet-2 using the full PSF, and ($vi$) the IsoNet-2 using the split PSF. In addition to the visuals given in the figure, Table 1 compares the PSNR of the full volumes with the two ground truth versions. As can be seen, our method performs best in all cases. Note that the failing to incorporate the PSF (as with pure SRCNN) results in an inferior reconstruction.

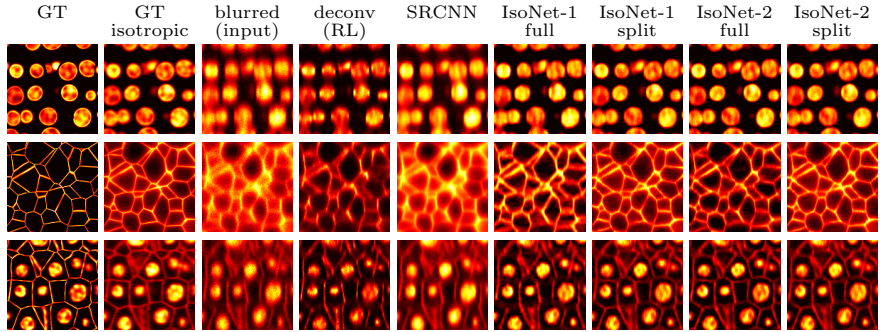

Fig. 2: Comparison of results on synthetic data. Rows show axial slices of 3D nuclei data, membrane data, and a combined dataset, respectively. The columns are: (*i*) ground truth phantom fluorophore densities, (*ii*) the same ground truth convolved with an isotropic PSF, (*iii*) anisotropically blurred isotropic GT image (the input images to all remaining columns, (*iv*) deconvolved images using Richardson-Lucy [13, 10], (*v*) SRCNN [1], (*vi*) IsoNet-1 with one (full) PSF, (*vii*) IsoNet-1 making use of the split PSFs, (*viii/ix*) IsoNet-2 with full PSF and split PSFs, respectively.

| volume (PSF/scale) | blurred (input) | deconv (RL) | SRCNN | IsoNet-1 | | IsoNet-2 | |
|---|---|---|---|---|---|---|---|
| | | | | full | split | full | split |
| nuclei (gaussian/8) | 25.28 | 26.98 | 25.41 | 31.24 | 31.58 | 33.99 | **34.60** |
| | 23.19 | 27.72 | 26.27 | 29.35 | 29.21 | **30.16** | 29.74 |
| membranes (confocal/4) | 22.13 | 17.58 | 21.98 | 19.84 | 26.51 | 19.45 | **27.67** |
| | 15.98 | 30.14 | 29.04 | 30.05 | 29.42 | **30.26** | 29.28 |
| nuclei+memb. (light-sheet/6) | 27.91 | 24.29 | 28.47 | 25.33 | 30.00 | 25.15 | **30.72** |
| | 24.25 | 26.78 | 24.64 | 26.96 | 26.15 | **27.71** | 26.57 |

Table 1: Computed PSNR values against isotropic GT (upper rows), and against GT (lower rows). PSF types are: gaussian ($\sigma_{xy}/\sigma_z = 2/8$); confocal with numerical aperture NA = 1.1; light-sheet with $\text{NA}_{\text{detect}} = 0.8$ and $\text{NA}_{\text{illum}} = 0.1$.

**Simple 3D Segmentation** To give a simple example of how the improved image quality helps downstream processing we applied a standard 3D segmentation pipeline on the simulated nuclei data (cf. Fig. 2). The segmentation pipeline con-sists of 3 simple steps: First, we apply a global threshold that is calculated using the *intermodes* method [15]. Then, holes in thresholded image regions are closed. Finally, cells that clump together are separated by applying a 3D watershed algorithm on the euclidian distance transform (computed in 3D). This pipeline is freely available to a large audience in tools like Fiji or KNIME. We applied this pipeline to the isometric ground truth data, the blurred and subsampled input data, and the result produced by the *IsoNet-2* . The final segmentation results in Table 2 demonstrate the effectiveness of the *IsoNet-2* in facilitating downstream segmentation.

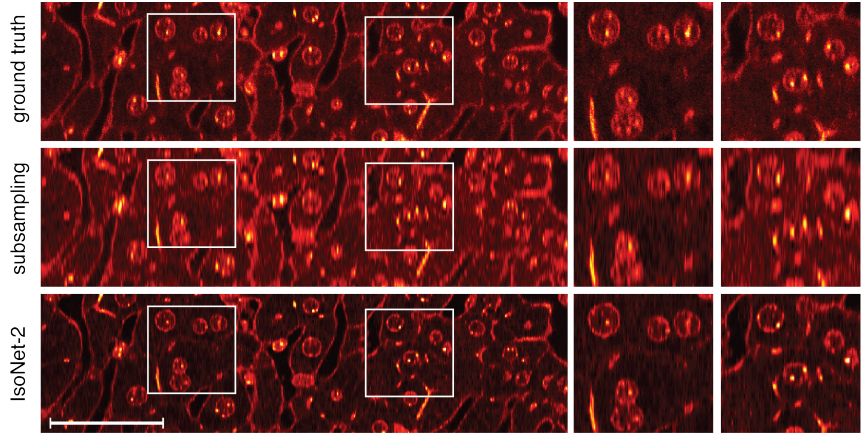

Fig. 3: Our results on fluorescence microscopy images of liver tissue (data taken from [11]). Nuclei (DAPI) and membrane (Phalloidin) staining of hepatocytes, imaged with a two-photon confocal microscope (excitation wavelength 780nm, NA=1.3, oil immersion, n=1.49). We start from an isotropic acquisition (ground truth), simulate an anisotropic acquisition (by taking every 8th slice), and compare the isotropic image to the IsoNet-2 recovered image. Scalebar is $50\mu m$.

|  | isotropic (GT) | anisotropic | IsoNet-2 |
| --- | --- | --- | --- |
| SEG | 0.923359 | 0.741533 | 0.912790 |

Table 2: Segmentation results on synthetic nuclei data. Evaluation metric is SEG (ISBI Tracking Challenge in 2013), the average intersection over union of matching cells when compared to the ground truth labels. SEG takes values in $[0, 1]$, where 1 corresponds to a perfect voxel-wise matching.

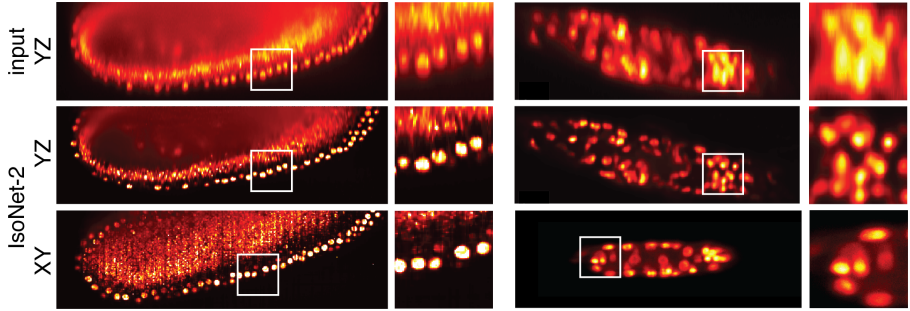

Fig. 4: IsoNet-2 applied to *Drosophila* (left) and *C. elegans* (right) volumetric images. We train on XY slices, and then apply the trained IsoNet-2 on the raw YZ input slices (upper row). The image quality of the recovered IsoNet-2 YZ slices (middle row) is significantly improved and shows similar isotropic resolution when compared to the XY slices (lower row).

### 3.2 Real Data

Furthermore, we validate our approach on confocal and light-sheet microscopy data and demonstrate the perceptual isotropy of the recovered stacks.

First we show that artificially subsampled two-photon confocal acquisitions can be made isotropic using *IsoNet-2*. As can be seen in Figure 3 the original isotropic data is nearly perfectly recovered from the 8-fold subsampled data (by taking every 8th axial slice). Second, we show that single view light-sheet acquisitions can be made isotropic. Fig. 4 shows stacks from two different sample recordings where we trained *IsoNet-2* to restore the raw XZ slices. The final results exhibit perceptual sharpness close to that of the higher quality raw XY slices, demonstrating the ability to restore isotropic resolution from a single volume in different experimental settings.

## 4 Discussion

We presented a method to enhance the axial resolution in volumetric microscopy images by reconstructing isotropic 3D data from non-isotropic acquisitions wit convolutional neural networks. This can be understood as restoring isotropy by deconvolving the image under the trained, sample specific image prior. Training is performed unsupervised and end-to-end, on the same anisotropic image data for which we recover isotropy. We have showed results on 3 synthetic and 3 real datasets and compared our results to the ones from Richardson-Lucy deconvolution [13, 10] and state-of-the-art super resolution methods. We finally further showed that a standard 3D segmentation pipeline performed on outputs of IsoNet-2 are essentially as good as on full isotropic data.

It seems apparent that approaches like the ones we suggest bear a huge potential to make microscopic data acquisition significantly more efficient. For the liver data, for example, we show (Figure 3) that only 12.5% of the data yields isotropic reconstructions that appear on par with the full isotropic volumes. This would potentially reduce memory and time requirements as well as laser induced fluorophore and sample damage by the same factor. Still, this method can, of course, not fill in missing information: If axial sample rate would drop below the Shannon limit (with respect to the smallest structures we are interested in resolving), the proposed networks will not be able to recover the data.

**Acknowledgments** We thank Valia Stamataki and Christopher Schmied (Tomancak lab) for the *Drosophila* dataset, Stephanie Merret and Stephan Janosch (Sarov Group) for the *C.elegans* images, Hernan Andres Morales Navarrete (Zerial lab) for the liver dataset, and Uwe Schmidt (all MPI-CBG) for helpfull feedback.